\documentclass[11pt,a4paper]{article}
\usepackage[hyperref]{emnlp2020}
\usepackage{times}
\usepackage{latexsym}
\usepackage{listings}
\usepackage[frozencache,cachedir=.]{minted}
\usepackage[T1]{fontenc}
\usepackage[whole]{bxcjkjatype}

\usepackage{microtype}

\aclfinalcopy 

\title{fugashi, a Tool for Tokenizing Japanese in Python}

\author{Paul McCann \\
  Cotonoha \\
  \texttt{howdy@cotonoha.io} }

\date{}

\begin{document}
\maketitle
\begin{abstract}
Recent years have seen an increase in the number of large-scale multilingual NLP projects. However, even in such projects, languages with special processing requirements are often excluded. One such language is Japanese. Japanese is written without spaces, tokenization is non-trivial, and while high quality open source tokenizers exist they can be hard to use and lack English documentation. This paper introduces fugashi, a MeCab wrapper for Python, and gives an introduction to tokenizing Japanese.
\end{abstract}

\section{Introduction}

Over the past several years there's been a welcome trend in NLP projects to be
broadly multilingual. However, even when many languages are supported, there are 
a few that tend to be left out. One of these is Japanese. Japanese is written 
without spaces, and deciding where one word ends and another begins is not
trivial. While highly accurate tokenizers are available, they can be hard to 
use, and English documentation is scarce. This is a short guide to tokenizing 
Japanese in Python that should be enough to get you started adding Japanese 
support to your application. \footnote{This paper was originally made available as a blog post. \url{https://www.dampfkraft.com/nlp/how-to-tokenize-japanese.html}}

This paper will begin with a tutorial on Japanese tokenization using fugashi, along with notes on issues to be aware of. Following that will be a discussion of the development of fugashi, closing with a brief overview of other Japanese tokenizers usable in Python.

\begin{table*}[]
    \footnotesize
    \centering
    \begin{tabular}{llllllll}
    麩 & フ & フ & 麩 & 名詞-普通名詞-一般 &  &  & 0 \\
菓子 & カシ & カシ & 菓子 & 名詞-普通名詞-一般 &  &  & 1 \\
は & ワ & ハ & は & 助詞-係助詞 &  &  &  \\
、 &  &  & 、 & 補助記号-読点 &  &  &  \\
麩 & フ & フ & 麩 & 名詞-普通名詞-一般 &  &  & 0 \\
を & オ & ヲ & を & 助詞-格助詞 &  &  &  \\
主材 & シュザイ & シュザイ & 主材 & 名詞-普通名詞-一般 &  &  & 0 \\
料 & リョー & リョウ & 料 & 接尾辞-名詞的-一般 &  &  &  \\
と & ト & ト & と & 助詞-格助詞 &  &  &  \\
し & シ & スル & 為る & 動詞-非自立可能 & サ行変格 & 連用形-一般 & 0 \\
た & タ & タ & た & 助動詞 & 助動詞-タ & 連体形-一般 &  \\
日本 & ニッポン & ニッポン & 日本 & 名詞-固有名詞-地名-国 &  &  & 3 \\
の & ノ & ノ & の & 助詞-格助詞 &  &  &  \\
菓子 & カシ & カシ & 菓子 & 名詞-普通名詞-一般 &  &  & 1 \\
。 &  &  & 。 & 補助記号-句点 &  &  &  \\
EOS \\\hline

    \end{tabular}
    \caption{Example output from fugashi on the command line. Each column is a different field from UniDic. This format can be customized; the format here is the default format distributed with UniDic 2.1.2. ``EOS" means ``End of Sentence", though MeCab does not perform sentence tokenization, and EOS is simply emitted at the end of any output.}
    \label{tab:output}
\end{table*}

\section{Preparation}

First, you'll need to install fugashi and a tokenizer dictionary. For this tutorial we'll use fugashi with unidic-lite. You can install them with this command:

\begin{minted}{shell}
pip install fugashi[unidic-lite]
\end{minted}

fugashi comes with a script so you can test it out at the command line. Type in some Japanese and the output will have one word per line, along with other information like part of speech. Refer to Table~\ref{tab:output} for an example.\footnote{All examples in this text use fugashi v1.0.4 and unidic-lite v1.0.7.}

\section{Sample Code}

Now we're ready to get started with converting plain Japanese text into a list of words in Python.

\begin{minted}[fontsize=\footnotesize]{python}
import fugashi
# This is our sample text.
# "Fugashi" is a Japanese snack primarily made 
# of gluten.
text = "麩菓子は、麩を主材料とした日本の菓子。"

# The Tagger object holds state about the 
# dictionary. 
tagger = fugashi.Tagger()

words = [word.surface for word in tagger(text)]
print(*words)
# => 麩 菓子 は 、 麩 を 主材 料 と し た 日本 
# の 菓子 。
\end{minted}

This prints the original sentence with spaces inserted between words. In many cases, that's all you need, but fugashi provides a lot of other information, such as part of speech, lemmas, broad etymological category, pronunciation, and more. This information all comes from UniDic \cite{unidic}, a dictionary provided by the National Institute for Japanese Language and Linguistics (NINJAL).\footnote{ Besides the version for modern written Japanese used here, there are also versions of UniDic for spoken Japanese and different historical varieties of the language, all available from the UniDic homepage. \url{https://unidic.ninjal.ac.jp/}}

fugashi is a wrapper for MeCab \cite{mecab}, a C++ Japanese tokenizer. MeCab is doing all the hard work here, but fugashi wraps it to make it more Pythonic, easier to install, and to clarify some common error cases.

You may wonder why part of speech and other information is included by default. In the classical NLP pipeline for languages like English, tokenization is a separate step before part of speech tagging. In Japanese, however, knowing part of speech is important in getting tokenization right, so they're conventionally solved as a joint task. This is why Japanese tokenizers are often referred to as ``morphological analyzers" (形態素解析器 \textit{keitaisokaisekiki}).

\begin{table*}[]
    \footnotesize
    \setlength{\tabcolsep}{4pt}
    \renewcommand{\arraystretch}{1.2}
    \begin{flushleft}
    \begin{tabular}{l|llllllllllllllll}
    \textbf{Input} & 麩 & を & 用い & た & 菓子 & は & 江戸 & 時代 & から & すでに & 存在 & し & て & い & た & 。\\
    \textbf{Lemmas} & 麩 & を & 用いる & た & 菓子 & は & エド & 時代 & から & 既に & 存在 & 為る & て & 居る & た & 。 \\
    \textbf{Translation} & \multicolumn{16}{l}{Snacks using gluten already existed in the Edo Period.}\\\hline
    \end{tabular}
    \begin{tabular}{l|lllllll}
    \textbf{Input} & すもも & も & もも & も & もも & の & 内 \\
     \textbf{Lemmas} & 李 & も & 桃 & も & 桃 & の & 内 \\
     \textbf{Translation} & \multicolumn{7}{l}{Japanese plums and peaches are both kinds of peaches.}\\\hline
    \end{tabular}
    \begin{tabular}{l|ll}
    \textbf{Input} & 彷徨う & 陽射し \\
    \textbf{Lemmas} & さ迷う & 日差し \\
    \textbf{Translation} & \multicolumn{2}{l}{Wandering sunbeams.}\\\hline
    \end{tabular}
    \end{flushleft}
    \centering
    \caption{Input tokens and their associated lemmas. Lemmas may not bear any visual resemblance to the raw forms, which can look like an error to users unfamiliar with Japanese, and can be surprising even to Japanese speakers.}
    \label{tab:lemmas}
\end{table*}

\section{Notes on Japanese Tokenization}

There are several things about Japanese tokenization that may be surprising if you're used to languages like English.

\subsection{Lemmas May Not Resemble the Words in the Text at All}

Here's how you get lemma information with fugashi:

\begin{minted}[fontsize=\footnotesize]{python}
import fugashi
tagger = fugashi.Tagger()
text = "..."

print("input:", text)
for word in tagger(text):
    # feature is a named tuple 
    # holding all the Unidic info
    print(word.surface, word.feature.lemma, 
        sep="\t")
\end{minted}

For the output of the script refer to Table~\ref{tab:lemmas}.

You can see that 用い has 用いる as a lemma, and that し has 為る and い has 居る, handling both inflection and orthographic variation. すでに is not inflected, but the lemma uses the kanji form 既に.

An important detail here is that while MeCab provides all this information in its output, it's returned as unstructured text data. Conventionally a user could use MeCab's output formatting language to get just the fields they need, or output all fields and parse the output to get the desired fields. fugashi provides wrappers for UniDic formatted data that handle the parsing and put it in named tuples for structured access, like \texttt{word.feature.lemma}.

These lemmas come from UniDic, which by convention uses the ``dictionary form" of a word for lemmas. This is typically in kanji even if the word isn't usually written in kanji because the kanji form is considered less ambiguous. For example, この (\textit{kono}, ``this [thing]") has 此の (same pronunciation and meaning) as a lemma, even though normal modern writing would never use that form. This is also true of 為る in the above example.

This can be surprising if you aren't familiar with Japanese, but it's not a problem. It is worth keeping in mind if your application ever shows lemmas to your user for any reason, though, as it may not be in a form they expect.

Another thing to keep in mind is that most lemmas in Japanese deal with orthographic rather than inflectional variation. This orthographic variation is called 表記ゆれ \textit{hyoukiyure} and causes problems similar to spelling errors in English.

\subsection{Verbs Will Often Be Multiple Tokens}

Inflections of a verb will typically result in multiple tokens. This can also affect adjectives that inflect, like 赤い \textit{akai} (``red"). You can see this in the verbs at the end of the previous example, or see Table~\ref{tab:verbs}.

\begin{table*}
\centering
\begin{tabular}{lll}
\hline
\textbf{Raw Text} & \textbf{Tokenized Output} & \textbf{Translation} \\
\hline
見た & 見 | た & saw, looked \\
見ました & 見 | まし | た & saw, looked (polite) \\
見なかった & 見 | なかっ | た & did not see \\
受け渡した & 受け渡し | た & handed over \\
遊べませんでした & 遊べ | ませ | ん | でし | た & was unable to play (polite) \\ 
赤かった & 赤かっ | た & red (past tense) \\\hline
\end{tabular}

\caption{Examples of tokenized verbs and adjectives resulting in multiple tokens. Using UniDic results in fine-grained tokenization, where some tokens are not words in any conventional sense. One class of adjectives resembles verbs and will also frequently result in multiple tokens when inflected.}\label{tab:verbs}
\end{table*}

This would be like if ``looked" was tokenized into ``look" and ``ed" in English. This feels strange even to native Japanese speakers, but it's common to all modern tokenizers. The main reason for this is that verb inflections are extremely regular, so registering verb stems and verb parts separately in the dictionary makes dictionary maintenance easier and the tokenizer implementation simpler and faster. It also works better in the rare case an unknown verb shows up. (Verbs are a closed class in Japanese, which means new verbs aren't common.)

In the early 90s several tokenizers handled verb morphology directly, but that approach has been abandoned over time because of the advantages of the fine-grained approach \cite[pp.~21--22]{kudobook}. Depending on your application needs you can use some simple rules based on part of speech to lump verb parts together or just discard non-stem parts as stop words. 

\subsection{The Tagger Object Has a Startup Cost}

It's fast enough that you won't notice for one invocation, but creating the Tagger is a lot of work for the computer. When processing text in a loop it's important you re-use the Tagger rather than creating a new Tagger for each input.

Don't do this:

\begin{minted}[fontsize=\footnotesize]{python}
for text in texts:
    tagger = fugashi.Tagger()
    words = tagger(text)
\end{minted}

Do this instead:

\begin{minted}[fontsize=\footnotesize]{python}
tagger = fugashi.Tagger()
for text in texts:
    words = tagger(text)
\end{minted}

If you follow the second pattern MeCab shouldn't be a speed bottleneck for normal applications.

\subsection{Always Note Your Tokenizer Details}

If you publish a resource using tokenized Japanese text, always be careful to mention what tokenizer and what dictionary you used so your results can be replicated. Saying you used MeCab isn't enough information to reproduce your results, because there are many different dictionaries for MeCab that can give completely different results. Even if you specify the dictionary, it's critical that you specify the version too, since popular dictionaries like UniDic may be updated over time.\footnote{If for some reason you are unable to identify the version of your dictionary, at least report the number of entries it has, which can be used as a primitive checksum.}

\section{Development Background}

fugashi was originally developed as part of adding Japanese support to spaCy \cite{spacy} due to lack of maintenance of the mecab-python3\footnote{\url{https://github.com/SamuraiT/mecab-python3}} library, but has since evolved to differentiate itself from that library in a few ways. The primary goal of fugashi is to make it as easy as possible to get fast Japanese tokenization while improving access to existing linguistic resources. 

This section will introduce the important features of fugashi and touch on how they were implemented. These features are not unique in isolation, but bringing them together in one place is the distinguishing feature of fugashi.

\subsection{Binary Wheels}

``Wheels" are modern Python packages that can include platform-specific binary code.\footnote{PEP 427 -- The Wheel Binary Package Format 1.0 \url{https://www.python.org/dev/peps/pep-0427/}} Distributing wheels allows users to install compiled packages even without having a compiler or other necessary dependencies on their systems. fugashi provides wheels for Linux, OSX, and Windows, so that it can be installed with a single pip command. 

Before wheels for MeCab were provided, a user had to install it from source or through a package manager. Some Linux distributions like Debian use code that differs from the most recent source, making consistent use of MeCab difficult. Compiling MeCab on Windows is also known to be challenging. Providing wheels allows for consistent versioning and easy installs across platforms. This is critical for integration in open-source projects where maintainers want to support Japanese but don't have the time to set up a special development environment to handle it.

fugashi was the first MeCab wrapper to provide binary wheels for all of Windows, Linux, and OSX. The code used to build fugashi wheels was later used to distribute wheels for mecab-python3. 

\subsection{Dictionary Packages}

Use of MeCab requires a dictionary. Historically MeCab shipped with some dictionaries, but these have not been updated since 2013 (if not earlier). Installing dictionaries required manual configuration that could vary depending on how MeCab had been installed, which made integrating a dictionary in open-source Python packages difficult. 

As part of supporting fugashi, the UniDic and IPAdic dictionaries have been packaged so that they can be installed directly via pip.\footnote{The PyPI package names are \texttt{unidic}, \texttt{unidic-lite}, and \texttt{ipadic}.} IPAdic is a dictionary that, while not updated since roughly 2007, remains popular for natural language applications for a variety of reasons such as compatibility with historical benchmarks. The previously introduced UniDic is maintained by NINJAL and is the official dictionary of Japanese Universal Dependencies \cite{jaud}.

Because of size limitations on PyPI\footnote{PyPI is the Python Package Index, a service that hosts packages to be installed via pip. \url{https://pypi.org/}}, UniDic is provided in two flavors: the full UniDic package, based on the latest version, requires an extra download step, but is otherwise simple to install and configure. unidic-lite is based on the 2.1.2 release of UniDic, which is the most recent release to fit under PyPI's file size limit of 60MB compressed. Both of these dictionaries have been modified slightly to avoid issues with the default distribution such as marking unusual punctuation as nouns or tokenizing any numbers into individual digits.\footnote{The MeCab documentation provides instructions on how to mitigate some of these issues, but doesn't distribute modified dictionaries. \url{https://taku910.github.io/mecab/unk.html}}

Before fugashi was developed, mecab-python3 releases starting in 2018 included a bundled IPAdic. Leaving aside the issues with IPAdic being out of date, this approach is similar to Janome and makes the tokenizer easier to use, but has the downside that it makes it harder to use other dictionaries. In the case of mecab-python3 this also had the issue that it was a change from prior behavior without notice and caused some confusion. Following development of pip installable dictionaries for fugashi, the feature was backported to mecab-python3. 

\subsection{Structured Data}

Another important feature of fugashi is providing access to structured data. UniDic in particular provides a wealth of linguistic information, such as pronunciation, lemma, etymological category, pitch accent, and even foreign spelling.\footnote{"Foreign spelling" refers to the spelling of loanwords in the original language. For example, a naive romanization of ポール would be \textit{pooru}, but the UniDic lemma is \texttt{ポール-Paul}. Similarly パン \textit{pan}, "bread", has the lemma \texttt{パン-pao} because it comes from the Portuguese. These spellings can optionally be used in cutlet, a romanization tool based on fugashi. \url{https://github.com/polm/cutlet}} Traditionally this information would be presented in MeCab as a string and the application would parse it as necessary. By performing this parsing up-front and dealing with variations in dictionary format automatically, fugashi makes it more accessible for downstream applications.

This feature is common in tokenizers not based directly on MeCab, but fugashi was the first Python MeCab wrapper to include it, and is (to my knowledge) the only Python tokenizer providing structured access to all fields in UniDic.

\subsection{A Pythonic Interface}

Besides structured data, the changes fugashi makes to the MeCab API to make it more Pythonic are subtle but important. The most obvious example is that the \texttt{parseToNode} function, used to turn an input string into a Node object for each token, would normally return the head of a linked list. Navigating a linked list using member variables is unremarkable in C/C++ but distinctly odd in Python. In deference to the MeCab API mecab-python3 strictly maintains the old interface, while fugashi returns a Python list of nodes, allowing use in list comprehensions and other Pythonic idioms.

\subsection{Detailed Error Messages}

One other significant change is a creative workaround for failed initializations of the \texttt{Tagger} object. Issues like forgetting to install a dictionary are very common and show up as errrors at initialization, and are the most common cause of issues on mecab-python3's Github repository, but a bug in MeCab\footnote{\url{https://github.com/taku910/mecab/issues/57}} causes error messages to be unavailable when MeCab is used as a library. The workaround involves passing the intialization arguments to a separate class and getting the error message for that. This convoluted process is invisible to the user. This particular feature doesn't affect the API and has been backported to mecab-python3.

The text of the error message used when initialization fails is also a departure from MeCab's default error messages, which are all one-line and often leave users confused. In contrast the fugashi (or mecab-python3) error message includes a link to a detailed FAQ in the README, debug information, and a note that issues need not be filed in English. This was inspired by similarly detailed error messages in spaCy.\footnote{See the spaCy error code for examples of error messages written in a friendly style that include links to related issues or documentation. \url{https://github.com/explosion/spaCy/blob/master/spacy/errors.py}} 

\subsection{Speed}

The difference in processing speed between tokenizers can be dramatic. In developing fugashi I created a simple benchmark that counts words in Natsume Souseki's \textit{I Am a Cat} to make sure I wasn't unknowingly introducing performance issues. This is not reflective of all real-world workloads, but it is a good task for getting a rough idea of tokenizer speed. See Table~\ref{tab:comparison} for the results, which demonstrate that MeCab is very fast. The run times presented here are the average over ten runs. The source code for this benchmark is available online.\footnote{\url{https://github.com/polm/ja-tokenizer-benchmark}}

\begin{table}[]
    \footnotesize
    \centering
    \begin{tabular}{lrrl}
    \hline
    \textbf{Tokenizer} & \textbf{Time} & \textbf{Relative Time}  \\
    \hline
    mecab-python3 & 290 & 1.00 \\
    \textbf{fugashi} & 294 & 1.01 \\
    natto-py & 1173 & 4.04 \\
    kytea & 2254 & 7.77 \\
    sudachipy & 10103 & 34.83  \\
    janome & 16496 & 56.88  \\
    \hline
    \end{tabular}
    \caption{Processing time in milliseconds for a simple benchmark word count task. fugashi and mecab-python3 are roughly equivalent in speed, with other packages being slower.}
    \label{tab:comparison}
\end{table}

\section{Comparison with Other Tokenizers}

There are a tremendous number of tokenizers for Japanese, and a comprehensive comparison is beyond the scope of this paper. This is a short overview of other tokenizers usable in Python. 

Tokenizers usable in Python may be broadly grouped into three categories: MeCab wrappers, MeCab-like tokenizers, and other tokenizers. 

\subsection{MeCab Wrappers}

Over the years there have been many MeCab wrappers for Python, though only a few are still maintained. The original MeCab code\footnote{\url{https://github.com/taku910/mecab}} includes a SWIG\footnote{"Simplified Wrapper and Interface Generator". SWIG allows a developer to write an interface file for C/C++ code and generate wrappers in a variety of languages. \url{http://www.swig.org/}} wrapper which has been the basis of several tokenizers. The MeCab wrappers are the fastest Python tokenizers. 

\textbf{mecab-python3} is a MeCab wrapper based on the SWIG code included in the main MeCab repository, and is the oldest of the tokenizer packages mentioned here, with its first release in 2014. fugashi was initially developed in response to a lack of maintenance of mecab-python3, but since then I have taken over the project and maintain it in parallel with fugashi. Several developments in fugashi are based on personal pain points with mecab-python3, and improvements to fugashi that don't affect the API, like pip-installable dictionary support, have been backported. Because mecab-python3 is widely used the main priority of maintenance is keeping the existing API stable for legacy applications, while fugashi is free to make the API more Pythonic for use in new applications.

The \textbf{mecab}\footnote{\url{https://pypi.org/project/mecab/}} project on PyPI, formerly known as \textbf{mecab-python-windows}, is based on the same SWIG code as mecab-python3 and has basically the same API. It provided Windows wheels long before mecab-python3, but since mecab-python3 began offering Windows wheels the differences between the packages are relatively minor.

\textbf{natto-py}\footnote{\url{https://github.com/buruzaemon/natto-py}} is a MeCab wrapper that uses a cffi interface to avoid needing a compiler and has simplified some of the MeCab API to be more Pythonic. However, the cffi interface is slower than Cython or SWIG, and since a separate MeCab install with dictionary is required that still leaves the user responsible for getting configuration right.

\subsection{MeCab-like Tokenizers}

Some tokenizers more or less explicity copy the design of MeCab while adding features or improving usability. These tokenizers all started life as Python projects, which greatly simplifies tooling, but comes at the expense of speed; they are much slower than the MeCab wrappers. This is still fast enough for small to medium sized corpora, but presents issues when processing larger amounts of text.

The main features that make a tokenizer MeCab-like are the use of an extensive dictionary with part of speech information, typically accessed via a double-array trie, and use of the Viterbi algorithm to find a minimum cost tokenization of a string.

\textbf{Janome}\footnote{\url{https://github.com/mocobeta/janome}} is a pure Python tokenizer with a long history. It includes a slightly modified IPAdic with the addition of 令和 \textit{Reiwa}, the current era name. Since everything necessary is included it's very easy to use, and was an inspiration in the development of fugashi. However, since the implementation is in pure Python, it's much slower than MeCab; the Japanese FAQ says it's roughly ten times slower.\footnote{\url{https://mocobeta.github.io/janome/}} Because IPAdic is tightly integrated it's also not straightforward to use significantly different dictionaries, though there is experimental support for the IPAdic-based Neologd\footnote{\url{https://mocobeta.github.io/janome/\#experimental-neologd-v0-3-3}}.

\textbf{SudachiPy}\footnote{\url{https://github.com/WorksApplications/SudachiPy}} is a Python port of the Java-based Sudachi tokenizer \cite{sudachi}. It has a high-quality UniDic-like dictionary and multiple modes of segmentation, and has recently been used when creating gold corpora for NER datasets. I have contributed to the code base in the interest of improving performance, Cythonizing performance critical parts of the code, but unfortunately it is still significantly slower than MeCab. Like fugashi it distributes dictionaries as pip packages.

\subsection{Other Tokenizers}

Some tokenizers use a very different strategy than MeCab when tokenizing. The examples listed here all use a model to decide whether to treat each character boundary as a word boundary or not. 

\textbf{Nagisa}\footnote{\url{https://github.com/taishi-i/nagisa}} is a relatively new tokenizer implemented in Python and based on neural networks. It's easy to use, but at present is significantly slower than SudachiPy or Janome. 

\textbf{Juman++}\footnote{\url{http://nlp.ist.i.kyoto-u.ac.jp/EN/index.php?JUMAN++}} is implemented in C++ and uses neural networks to determine word boundaries \cite{jumanpp}. It has an official Python wrapper, but requires the core tokenizer to be installed separately, making configuration difficult. Version 2 of the software has had release candidates released annually since roughly 2017, but it's unclear which version should be used now. I attempted to do a simple benchmark using the most recent v2 release candidate but it failed with an error.  

\textbf{KyTea}\footnote{\url{http://www.phontron.com/kytea/}} uses logistic regression or SVM to determine word boundaries \cite{kytea}. It's implemented in C++, but it has a few Python wrappers. None of them distribute wheels, so it's necessary to install the C++ tokenizer on your own. It is slower than MeCab but faster than the MeCab-like tokenizers.

\section{Summary}

fugashi combines the speed of MeCab with the ease-of-use of more recent tokenizers, striking a balance that's widely useful. fugashi is not faster than existing tokenizers; it does not have a new or better dictionary; it does not have new features; it merely takes the best of the available resources, puts them together, and makes sure that everything works in a variety of environments with a minimum of effort. As noted in \cite{risk}, to document best practices is good, but to automate them is better.  While it's hoped that newer tokenizers like SudachiPy will be able to catch up in performance soon, at present fugashi is a good choice for many applications.

The past year has seen many new developments in the world of Japanese tokenizers. For more information on current Japanese tokenizers in Python, refer to Konoha\footnote{\url{https://github.com/himkt/konoha}} or Toiro\footnote{\url{https://github.com/taishi-i/toiro}}, which wrap multiple tokenizers and allow comparisons between them.

\section{Acknowledgments}

The author would like to thank all the contributors to fugashi, particularly Aki Ariga for providing Windows support.

\bibliography{anthology,emnlp2020}

\begin{thebibliography}{9}
\expandafter\ifx\csname natexlab\endcsname\relax\def\natexlab#1{#1}\fi

\bibitem[{Agirre et~al.(2018)Agirre, L{\'o}pez~de Lacalle, and Soroa}]{risk}
Eneko Agirre, Oier L{\'o}pez~de Lacalle, and Aitor Soroa. 2018.
\newblock \href {https://doi.org/10.18653/v1/W18-2505} {The risk of sub-optimal
  use of open source {NLP} software: {UKB} is inadvertently state-of-the-art in
  knowledge-based {WSD}}.
\newblock In \emph{Proceedings of Workshop for {NLP} Open Source Software
  ({NLP}-{OSS})}, pages 29--33, Melbourne, Australia. Association for
  Computational Linguistics.

\bibitem[{Asahara et~al.(2018)Asahara, Kanayama, Tanaka, Miyao, Uematsu, Mori,
  Matsumoto, Omura, and Murawaki}]{jaud}
Masayuki Asahara, Hiroshi Kanayama, Takaaki Tanaka, Yusuke Miyao, Sumire
  Uematsu, Shinsuke Mori, Yuji Matsumoto, Mai Omura, and Yugo Murawaki. 2018.
\newblock Universal dependencies version 2 for japanese.
\newblock In \emph{LREC}.

\bibitem[{Den et~al.(2008)Den, Nakamura, Ogiso, and Ogura}]{unidic}
Yasuharu Den, Junpei Nakamura, Toshinobu Ogiso, and Hideki Ogura. 2008.
\newblock A proper approach to japanese morphological analysis: Dictionary,
  model, and evaluation.
\newblock In \emph{LREC}.

\bibitem[{Honnibal and Montani(2017)}]{spacy}
Matthew Honnibal and Ines Montani. 2017.
\newblock {spaCy 2}: Natural language understanding with {B}loom embeddings,
  convolutional neural networks and incremental parsing.
\newblock To appear.

\bibitem[{Kudo(2018)}]{kudobook}
Taku Kudo. 2018.
\newblock \emph{形態素解析の理論と実装[Morphological Analysis:
  Theory and Implementation] (Japanese)}.
\newblock 近代科学社.

\bibitem[{Kudo et~al.(2004)Kudo, Yamamoto, and Matsumoto}]{mecab}
Taku Kudo, Kaoru Yamamoto, and Y.~Matsumoto. 2004.
\newblock Applying conditional random fields to japanese morphological
  analysis.
\newblock In \emph{EMNLP}.

\bibitem[{Morita et~al.(2015)Morita, Kawahara, and Kurohashi}]{jumanpp}
Hajime Morita, D.~Kawahara, and S.~Kurohashi. 2015.
\newblock Morphological analysis for unsegmented languages using recurrent
  neural network language model.
\newblock In \emph{EMNLP}.

\bibitem[{Neubig et~al.(2011)Neubig, Nakata, and Mori}]{kytea}
G.~Neubig, Yosuke Nakata, and S.~Mori. 2011.
\newblock Pointwise prediction for robust, adaptable japanese morphological
  analysis.
\newblock In \emph{ACL}.

\bibitem[{Takaoka et~al.(2018)Takaoka, Hisamoto, Kawahara, Sakamoto, Uchida,
  and Matsumoto}]{sudachi}
Kazuma Takaoka, Sorami Hisamoto, Noriko Kawahara, Miho Sakamoto, Yoshitaka
  Uchida, and Yuji Matsumoto. 2018.
\newblock \href {https://www.aclweb.org/anthology/L18-1355} {{S}udachi: a
  {J}apanese tokenizer for business}.
\newblock In \emph{Proceedings of the Eleventh International Conference on
  Language Resources and Evaluation ({LREC} 2018)}, Miyazaki, Japan. European
  Language Resources Association (ELRA).

\end{thebibliography}
\bibliographystyle{acl_natbib}

\end{document}